\documentclass[eng]{ajceam-class}
\usepackage{algorithm}
\usepackage{algpseudocode}

\usepackage[square,super,numbers,]{natbib}
\title{Comparative Analysis of Clustering Techniques for Personalized Food Kit Distribution }

\shorttitle{Personalization of Food kit using clustering}

\author[1]{Jude Francis}
\author[2]{Rowan K Baby}
\author[2]{Jacob Abraham}
\author[1]{Ajmal P.S}

\affil[1]{College of Engineering Trivandrum, Dept. of Mechanical Engineering, Kerala, India}
\affil[2]{College of Engineering Trivandrum, Dept. of Computer Science and Engineering, Kerala, India}

\firstauthor{Jude, Rowan, Jacob, and Ajmal}

\contactauthor{Jude Francis}           % Name and surname of the contact author
\email{jude.fjohn@gmail.com}            % Contact Author Email

\abstract{
The Government of Kerala had increased the frequency of supply of free food kits owing to the pandemic, however, these items were static and not indicative of the personal preferences of the consumers. This paper conducts a comparative analysis of various clustering techniques on a scaled-down version of a real-world dataset obtained through a conjoint analysis-based survey. Clustering carried out by centroid-based methods such as k means is analyzed and the results are plotted along with SVD, and finally, a conclusion is reached as to which among the two is better. Once the clusters have been formulated, commodities are also decided upon for each cluster. Also, clustering is further enhanced by reassignment, based on a specific cluster loss threshold. Thus, the most efficacious clustering technique for designing a food kit tailored to the needs of individuals is finally obtained.}


\begin{document}

% Include title, authors, abstract, etc.
\maketitle

\printcontactdata

% Main body of manuscript
\section{Introduction}
\firstword{T}{he}% Capital letter in first word
     most delicate sections of society often comprise individuals or families ridden with malnutrition, hunger, and those that are inextricably entangled within the vicious circle of poverty\cite{verma2021socio}. Hence, most governing authorities employ poverty thresholds to assess whether subsidies have to be sanctioned to such deserving candidates.\cite{arora2013food}    
    And one state that is spearheading the emancipation of individuals below the poverty line in India, is Kerala \cite{masiero2015redesigning}.
    
    With one of the highest literacy rates in the entire country- Kerala \cite{rathore2019explaining} had recently  caught the eye of the global stage through the exceptional way in which it had managed the challenges imposed by the first wave of the pandemic through certain ingenious strategies \cite{sulaiman2020trace}. And amidst the plethora of  health and safety-related actions taken, one such commendable effort was the government's decision to provide kits containing necessary food items, free of cost, \cite{Foodkits} to all ration card holders. Herein a solution has been devised to improve the disbursement of provisions in a more effective manner by employing a specific type of clustering algorithm that suits the task at hand.

\section{MOTIVATION}
    Prior to the pandemic, the  food distribution system was solely focused on providing a means to establish a steady and reliant channel of the food supply through government-intervened subsidization. The rate of subsidies was classified into various categories according to the financial capability of the individuals in each category \cite{ravi2019impact}. Thus, the underprivileged, who were the major beneficiaries of this system could easily access sufficient food. 
    
    However, food kits containing additional commodities were mostly given only during traditional festive seasons. And when the livelihoods of people were threatened due to the advent of COVID, the government increased the frequency of kits supplied, thereby ensuring that they would act as a safeguard for the common mass, assuring  them that their basic necessities would be met.
    
    However, the problem with the implementation of such a system was that the commodities included in these kits were not decided upon by carrying out any statistical analysis, nor were there any mechanisms set into  motion, to dynamically alter its constituents according to the tastes and trend patterns of its consumers. 
    
    Moreover, another inherent problem was that the consumption data of such food kits could never reflect  the innate preferences of its recipients, as they would rather accept food kits offered freely than simply refuse them. Hence, the administration could do nothing but impose a set of items that were commonly thought of as necessary, without any underlying studies or analysis to back up their decisions. Thus, the food kits offered were not able to satisfy the demands of the people, and that is why a new solution had to be devised in order to address this issue. 
    
\section{Problem Statement}

    In order to bridge the gap between the consumer and the food kit facilitator, one needs to ensure that the choices made by the facilitator align with the preferences of the consumer. Though it may seem that such a task could be easily carried out by just asking the consumer directly for his preference, the difficulty in actualizing this is mainly due to the following reasons: Firstly, it is impossible to allocate a unique kit to each individual, as it would lead to innumerable combinations. Thus, it must be conceded that only a limited number of kits can be allocated, which inevitably causes a mismatch between the interests of the individual and possible kit combinations. Secondly, the preferences of the consumers alone would not yield a result capable of solving this crisis, as the data collected would be spread out depending on multiple factors ranging from geographical to climatic conditions. Therefore, in order to accommodate the multifarious nature of the consumption pattern of various communities, one would inevitably conclude that a personalized kit should be designed from an optimal number of combinations. And once this has been fixed the next question would be to decide upon the level of personalization that the government has to adopt, which is not only viable from a theoretical but also an economical viewpoint.

     So in this study, a system has been proposed to address the aforementioned problems, by carrying out a replica of the scaled-down version of the whole situation using a sampled population, and analyzing how various methods fare in providing satisfactory results.

\section{Solution Approach}
    \subsection{Constraints}
        \begin{figure}[htp]
            \includegraphics[scale=0.4]{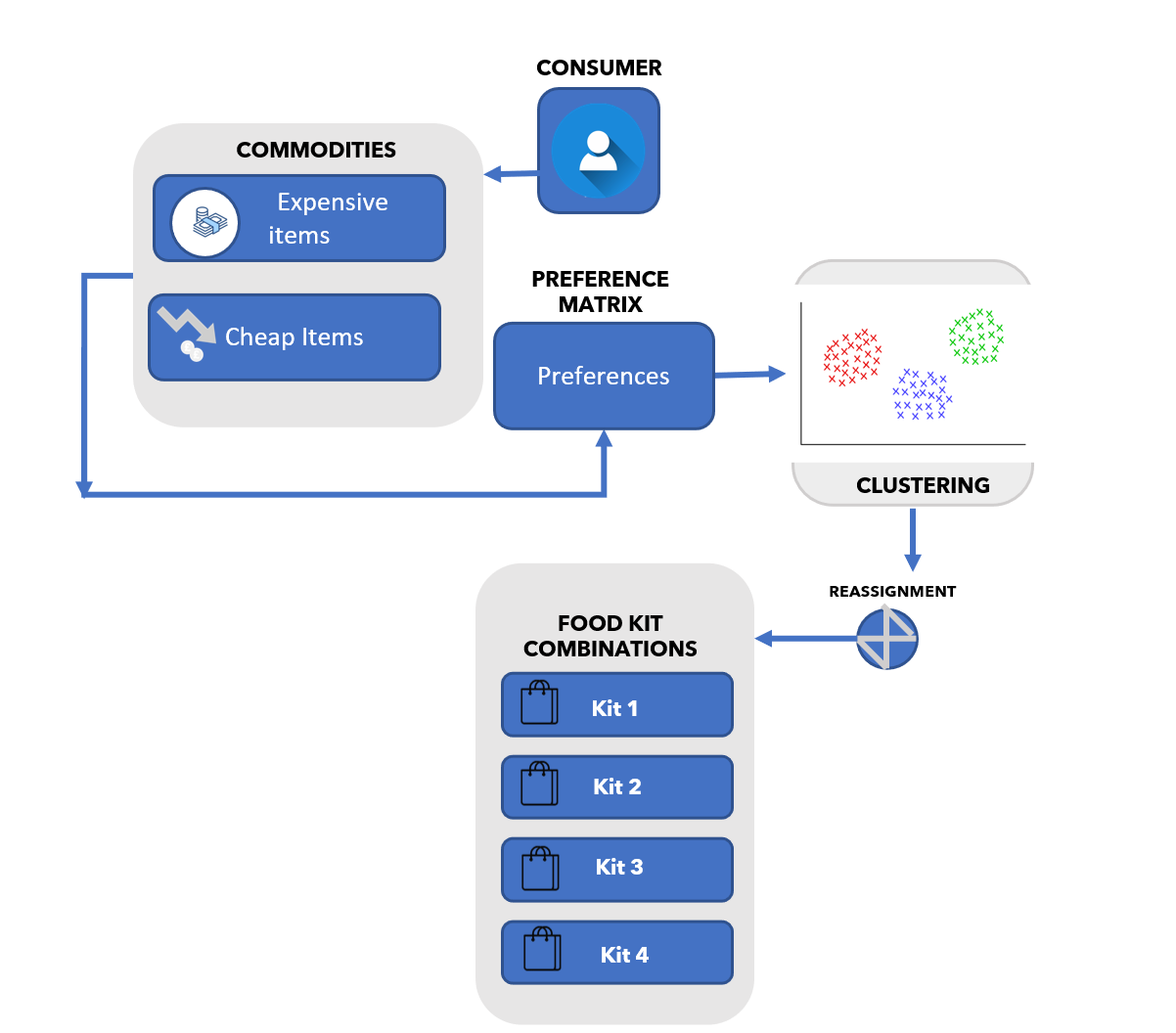}
            \caption{Solution Flowchart}
        \end{figure}
        The first stage of the solution was to assess the preference of individuals so as to ensure that the decisions taken were actually fine. So a random convenience sampling of about 200 people from Kerala was taken, who willingly volunteered for a choice-based conjoint analysis of various potential commodities that the government was willing to offer. The conjoint analysis would effectively help in gauging the importance of items based on psychological trade-offs taken by the consumer in order to maximize the net benefits obtained. \par
        The choices for the scaled-down study were kept at 20 essential items out of which these items were further added to two categories expensive and cheap, in order to prevent people from selecting just the expensive items, rather than what they actually needed. The participants were allowed to select only 6 items from the expensive category and 4 from the cheap category, resulting in a total of 10 items. 
        
        After the data collection was carried out, two choices had to be made whether an item-based approach or a user-based approach had to be adopted.
    \subsection{Clustering Algorithms}
        The various types of clustering algorithms are as follows:

        \textbf{Hierarchical Clustering}
            \begin{itemize}
                \item  Hierarchical clustering creates a tree of clusters. 
               \item It is well suited to hierarchical data, such as taxonomies.
                \item Any number of clusters can be chosen by cutting the tree at the right level.
            \end{itemize}

        \textbf{Centroid-based Clustering}
            \begin{itemize}
                \item  Centroid-based clustering organizes the data into non-hierarchical clusters, in contrast to hierarchical clustering. k-means is the most widely-used centroid-based clustering algorithm.
                \item Centroid-based algorithms are efficient but sensitive to initial conditions and outliers
            \end{itemize}

        \textbf{Density-based Clustering}
            \begin{itemize}
                \item Density-based clustering connects areas of high example density into clusters. This allows for arbitrary-shaped distributions as long as dense areas can be connected. 
                \item These algorithms have difficulty with data of varying densities and high dimensions. Further, by design, these algorithms do not assign outliers to clusters.
            \end{itemize}

        \textbf{Distribution-based Clustering}
            \begin{itemize}
                \item  This clustering approach assumes data is composed of distributions, such as Gaussian distributions.
                \item  As the distance from the distribution's center increases, the probability that a point belongs to the distribution decreases.
            \end{itemize}
         And for this problem statement, since the requirement was not to form any hierarchical patterns, partition-based methods were the next best choice. Hence,k-means (a centroid-based algorithm ) was chosen as the appropriate clustering algorithm

\section{Algorithm used}

    \subsection{K-means algorithm}
        The K- means algorithm has gained quite a popularity over the years owing to its ease and simplicity of implementation, scalability, speed of convergence, and adaptability to sparse data. Basically, it can be thought of as a gradient descent procedure, which begins at starting cluster centroids, and iteratively updates these centroids to decrease the objective function. Some of the various merit measures for evaluating the techniques applied:
        \begin{enumerate}
            \item Graphical aid to the interpretation and validation of cluster analysis.
            \item The Silhouette width
        \end{enumerate}

	    Here, the\textbf{ k}-means algorithm is considered along with Euclidean distance to measure the similarity of the customer’s preference. The dataset is represented as n-points, with each point corresponding to whether a customer wishes a particular item to belong to the food kit. And the end goal is to ensure that the designed kits yield a maximum number of customers with a minimum level of dissatisfaction.

        To attain this, \textit{‘k’} clusters are considered. Their value is altered until a satisfactory value is  reached. \textit{’K’} points \((m_1,m_2,m_3,...m_k)\) have to be found out, that act as the cluster centroids and the points \((x_1,x_2,..x_n)\) belonging to any arbitrary cluster \textit{‘i’} will be such that the Euclidean distance amongst the points is always less than the distance with any point outside that particular cluster.

        \begin{algorithm}
        \caption{ K-Means}\label{alg:cap}
        \begin{algorithmic}
            \State $shuffleRow = randperm(size(input, 1));$
            \State $data = input(shuffleRow, :);$ \Comment{Randomize the data}
            \State $Y = data(:, 2:end);$
            \State $K = 15;$ \Comment{Centroid limit}
            \State $lambda = 0.3;$ \Comment{Learning Rate}
            \While{$iter \neq K$}
            \State $initial\_centroids = Y(1:iter, :);$
            \State $max\_iters = 100;$
            \State $idx = findClosestCentroids(Y,initial\_centroids, lambda);$
            \State $centroids = computeCentroids(Y, idx, iter);$
            \State $runkMeans(Y, initial\_centroids, max\_iters);$
            \State $bestCentroid(Y, idx, iter);$ \Comment{Find the best kits}
            \State $silhouette(Y, centroids, idx, iter);$ \Comment{Evaluate Error}
            \State $iter = iter + 1 $ \Comment{Evaluate Error}
            \EndWhile
        \end{algorithmic}
        \end{algorithm}

        For Algorithm 1 defined it takes input data. The input will contain a list of users with their corresponding preferences for the items available for the kit. The user is required to choose 10 items out of a list of m items. In the algorithm, the value of K can vary according to the needs and is analyzed as shown in the next section. The value of lambda will range from 0 to 1. The most suitable value after various trials was found to be 0.3 and it has been used here. Similarly, after various trials, it was found that the value for maximum iterations (max\_iter) should have a value of 100. The methods to compute the closest centroid, running the K-Means, and finding the best centroid corresponds to the standard algorithm for K-Means.

    \subsubsection{\textbf{Result}}

    \begin{table}[htp]
    \begin{tabular}{|c||c||c||c|}
         \hline
            Kits & \multicolumn{3}{|c|}{Trials}\\
          \hline
            4 & 0.128095 & 0.1612425 & 0.1611875\\
          \hline
            5 & 0.139614 & 0.168186 & 0.167112\\
          \hline
            6 & 0.1509433333 & 0.2312133333 & 0.1858016667\\
          \hline
            7 & 0.1504685714 &	0.22687 & 0.1879557143\\
          \hline
            8 & 0.15819375 & 0.21523 & 0.1954125\\
          \hline
            9 & 0.1823155556 & 0.2282477778 & 0.2058988889\\
          \hline
            10 & 0.189969 & 0.222137 & 0.196728\\
          \hline
            11 & 0.1881945455 &	0.1992954545 & 0.1760672727\\
          \hline
            12 & 0.2241175 & 0.2268658333 & 0.1781225\\
          \hline
            13 & 0.2242584615 &	0.2309376923 & 0.1892176923\\
          \hline
            14 & 0.2435464286 & 0.234595 & 0.2099157143\\
          \hline
            15 & 0.2665806667 &	0.237452 & 0.2598893333\\
          \hline
    \end{tabular}
    \caption{Silhouette width}
    \end{table}

    A preset condition of a minimum of 4 kits was already set as per the design constraint, hence the value of k<4 is not considered at all. This condition was set in order to ensure that the people will be provided with a minimum variety to choose from as the taste of the general public will vary depending on their geographical and cultural diversity.

    The given \textbf{Table 1} contains the plot for the various trials and the performance evaluation for each of the trials. This table has the first column as the number of kits and in the algorithm, it corresponds to the number of clusters. Each of the columns after that corresponds to three separate trials conducted using the K-Means algorithm. The values in these columns correspond to the Silhouette width, here the average silhouette width for the kits is taken for each iteration of the trial. This was done by taking the sum of the errors for each kit and dividing it by the total number of kits. \textbf{Fig 2} shows the performance of the K-Means in a graphical format with respect to the three trials given in Table 1.

    Collectively, from both the table and the graph, one reaches the conclusion that the results obtained from this algorithm are a bit ambiguous and can be clearly seen by the variation in the lines in the \textbf{Fig 2}. We can attribute this to an error in the data that is being used or it can be the insufficiency of the algorithm used. 
    To perform further analysis in this respect the number of kits considered in the trial is increased to a very large number.

    \begin{figure}[htp]
        \includegraphics[scale=0.42]{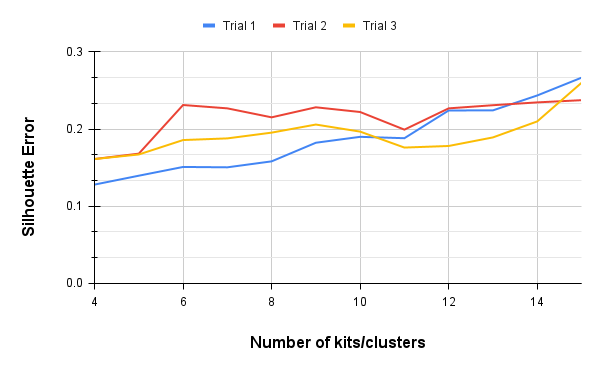}
        \caption{Performance of K-Means}
    \end{figure}

    \begin{figure}[htp]
        \includegraphics[scale=0.42]{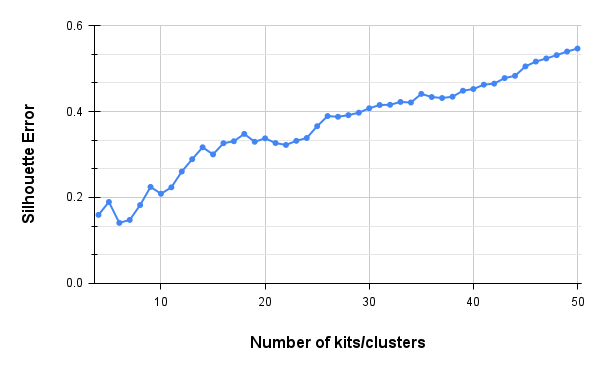}
        \caption{Performance of K-Means}
    \end{figure}

    A trial is run by taking 50 kits or clusters as the limit to the number of clusters to further prove or disprove the inconsistency of the data or the algorithm that has been used. Fig 2 shows the variation in the performance of the algorithm with respect to the number of clusters used. Here from the graph, the performance is varying in a consistent manner and this shows how the data used is consistent for this experiment.
    
    It is noticed that as the number of clusters drops, the errors formed also reduce, but such a conclusion is naive and simply not true. Moreover, the values of the number of users in each cluster keep on varying each time the trial is run, resulting in a highly erroneous and erratic output.

    To make a more conclusive and strong statement on the kits and the various contents, one has to resort to other clustering means, in order to obtain a more  stream-lined and convergent conclusion
    
\section{Singular Value Decomposition (SVD)}
    The singular value decomposition of a matrix \textbf{A}\( _{p\times q}\) with rank \textbf{r} is an orthogonal decomposition of a matrix into three matrices such that \(\textbf{A}_{p\times q} \textbf{ = } \textbf{U}_{p\times r}\textbf{\(\Sigma\)}_{r\times r}\textbf{V}^T_{r\times q}\). The SVD of a matrix is an exact decomposition, thus, the original matrix can be re-obtained by multiplying \( \textbf{U}\), \(\Sigma\), and \(\textbf{V}^T\)

    However, we can also approximate the given matrix \textbf{A} to a rank\textbf{ r} by multiplying the\textbf{ r} columns of U, the first r values of \(\sigma\) matrix and \textbf{r} rows of the \( \textbf{V}^T\) matrix, resulting in a truncated SVD matrix.

    In order to understand the application of \textbf{ SVD} intuitively related to the problem statement at hand, it is inevitable to understand what each of the resulting 3 matrices obtained from the eigen decomposition on the dataset procured through conjoint analysis implies, and it is stated as follows:

    \begin{enumerate}
        \item The left singular matrix denoted by \textbf{U} represents the user-latent space relationship matrix.
        \item The singular value matrix denoted by \textbf{\(\Sigma \)}  represents the extent of scaling.
        \item The right singular matrix denoted by \textbf{ \(V^T \)} represents the item-latent space relationship matrix
    \end{enumerate}

    And in terms of its eigen decomposition, they can be defined as follows:
    \begin{enumerate}
        \item  The columns of V (right-singular vectors) are eigenvectors of \(\textbf{A}^T\times\textbf{ A}\).
        \item  The columns of U (left-singular vectors) are eigenvectors of \(\textbf{A}\times \textbf{A}^T\).
        \item  The non-zero elements of \textbf{\(\Sigma\)} (non-zero singular values) are the square roots of the non-zero eigenvalues of \(\textbf{A}^T\times \textbf{A}\) or \(\textbf{A} \times \textbf{A}^T\)
    \end{enumerate}

    \begin{figure}
        \centering
        \includegraphics[scale=0.6]{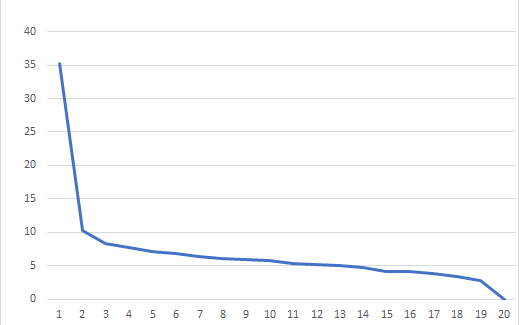}
        \caption{ Variation of  {\(\sigma\)} (Y axis) with \textbf{rank r} (X-axis)}
    \end{figure}

    The \textbf{\(\Sigma\)} matrix is a diagonal matrix with \(\sigma\) values  arranged in descending order along the leading diagonal, which gives a clear indication of the strength of the scaling. And on plotting these sigma values, it is seen from \textbf{figure 4}  that the difference in the consecutive values is significant only up to the range between 2- 4,  and thereafter the values do not change much.

    \subsection{The reason why the item-based approach does not work?}
        Clustering can be done either in an item-based manner or a user-based manner. The item-based clustering is done by the analysis of the first \textbf{r} rows of the \( \textbf{V}^T\) Matrix and then applying the SVD signs method, whereby the columns having the same sign patterns are clustered together.
        The results of such an item-based clustering are shown in \textbf{Fig.5} It is observed that as the rank increases, a larger number of items are included in that particular cluster. However, such a pattern is not logically correct, because the increase in the number of clusters points to a failure in effectively grouping the items. Hence, as seen from \textbf{Fig 5}, once the number of clusters reaches 17 out of the 20 available items each cluster would comprise mostly of one item, which is why an item-based approach is not suitable for this problem statement.

    \begin{figure}[htp]
        \centering
        \includegraphics[scale=0.6]{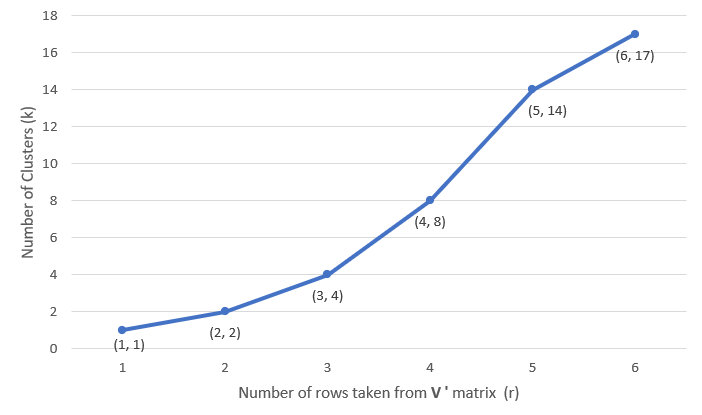}
        \caption{Variation of Number of Clusters with Rank r}
    \end{figure}
    \subsection{User-based approach}
        Hence, the second method - i.e. the user-based approach, where clustering of the users was carried out to obtain different smaller groups or clusters where the individual within each cluster would show similar characteristics and preferences in terms of the items chosen.

        The first \textbf{r} columns of the left singular matrix, representing the user-latent space matrix,  are considered and the rest are truncated, and using the \textit{SVD signs method} \cite{douglas2008clustering}, the corresponding clusters are obtained. As inferred from \textbf{fig 3}, the value for k is taken as 4 and the following steps were carried out to simplify the whole clustering process.
        
        \begin{enumerate}
            \item The positive values were replaced by 1 and the negative numbers were replaced by 0.
            \begin{enumerate}
                \item Thus, each row gave a binary representation corresponding to a unique number. The rows having the same binary forms would belong to the same cluster.
            \end{enumerate}
            \item Once the clusters have amalgamated, the next step was to design a kit for a particular cluster and that can  be done by counting the frequency of occurrence of each item in a particular cluster.
            \item The top 10 items will decide the items assigned to that particular kit.
            \begin{enumerate}
                \item Here, using the dataset, 8 clusters were obtained by taking the r-value as 4, and if r=2 were taken, then 16 clusters would've been obtained, and since that is highly uneconomical, we discarded the latter case and stuck to r=4 and the subsequent 8 clusters thus obtained.
            \end{enumerate}
            \begin{table}[htp]
            \begin{tabular}{ |p{3cm}||p{3cm}|  }
            \hline
             \hline
                Value of \textbf{r}& Number of clusters obtained\textbf{ (k) }\\
             \hline
                2 & 2\\
             \hline
                3 & 4\\
             \hline
                4 & 8\\
             \hline
                5 & 16\\
             \hline
                6 & 32\\
             \hline
            \end{tabular}
            \caption{}
            \end{table}
            \begin{figure}[htp]
                \centering
                \includegraphics[scale=0.6]{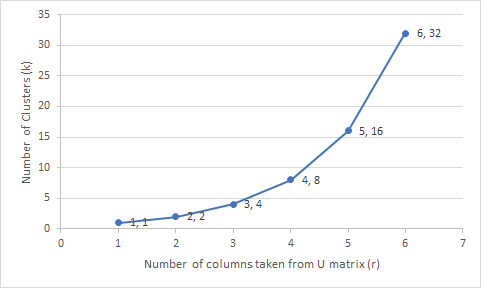}
                \caption{Variation of Number of Clusters with Rank r}
            \end{figure}
            \begin{figure}[htp]
                \centering
                \includegraphics[scale=0.6]{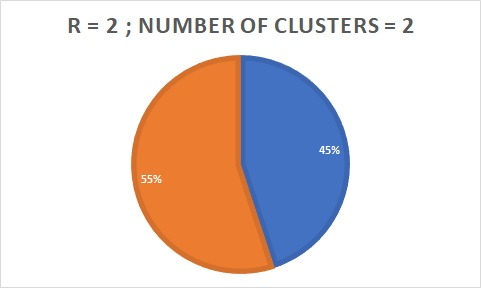}
                \caption{}
            \end{figure}
            \begin{figure}[htp]
                \centering
                \includegraphics[scale=0.6]{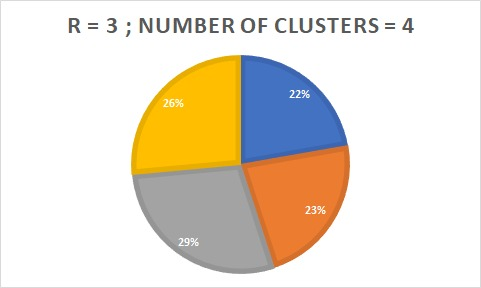}
                \caption{}
            \end{figure}
            \begin{figure}[htp]
                \centering
                \includegraphics[scale=0.6]{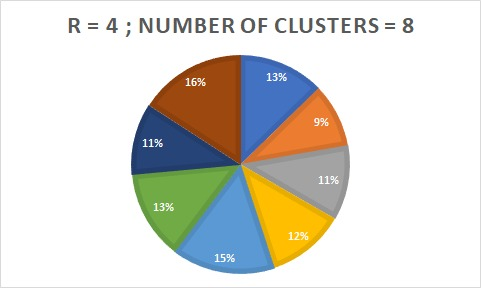}
                \caption{}
            \end{figure}
            \begin{figure}[htp]
                \centering
                \includegraphics[scale=0.6]{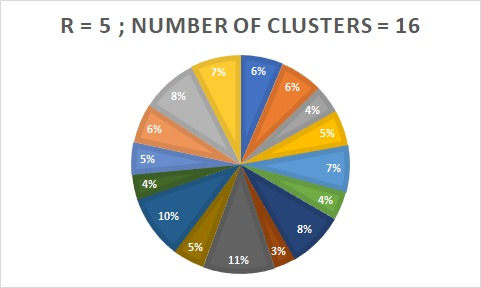}
                \caption{}
            \end{figure}
            \begin{figure}[htp]
                \centering
                \includegraphics[scale=0.6]{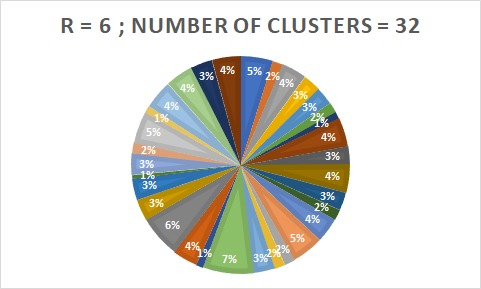}
                \caption{}
            \end{figure}
            \item Once the clusters and the kits have been obtained, the next step is to optimize the results obtained and a variety of metrics were employed to ensure that the solution obtained best suited the preferences of the individuals.
            \begin{enumerate}
                \item For the aforementioned reason, the items that each user preferred as marked in the initial survey analysis was  compared with the corresponding elements of each of the resultant 8  kits obtained, and for each kit, if these elements did not match, then a loss variable was defined to keep track of the number of mismatches and the kit having the lowest level of dissimilarity or error would be the kit to which the user would be assigned to. This is the main crux of the reassigning that is done to optimize the solution.
            \end{enumerate}
            \item As a final means of validation, the overall cluster loss of each cluster was calculated through two different means( normal and exponential average) to adjudge the efficacy of the reassignment process. The cluster loss of each individual before the reassignment process, as found out earlier, is again computed, but this time, the values obtained are averaged in order to compute the overall cluster loss of each cluster.(\textbf{fig 12} and \textbf{fig 13})                    
            \item Similarly, the overall cluster loss after reassignment is also computed,\textbf{fig 13} based on the exponential average amplifies the cluster losses of every cluster, and it can be inferred how the losses have decreased except for the second one. Thus by accepting a slight trade-off for the second cluster, the overall objective can be easily attained by carrying out the reassignment.
            \begin{figure}[htp]
                \centering
                \includegraphics[scale=0.55]{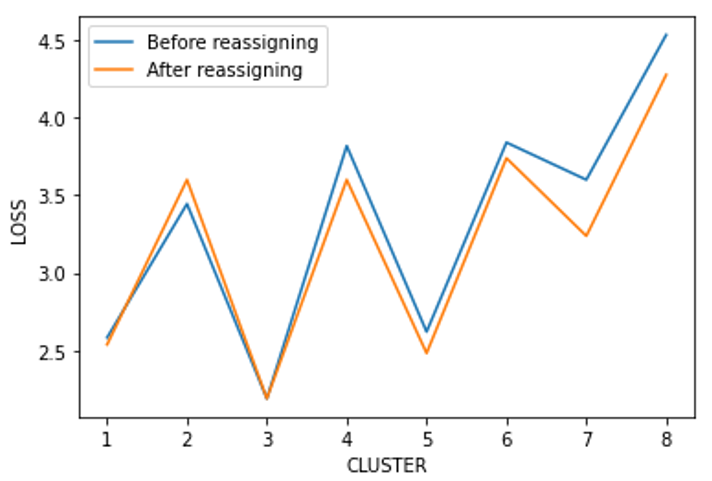}
                \caption{Normal average loss of cluster loss on reassignment}
            \end{figure}
            \begin{figure}[htp]
                \centering
                \includegraphics[scale=0.55]{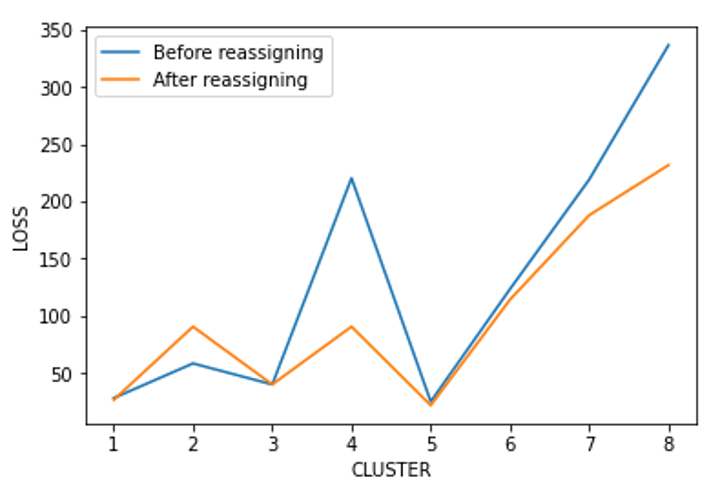}
                \caption{Exponential Averaging to magnify cluster loss on reassignment}
            \end{figure}
        \end{enumerate}

\section{Conclusion}
The food security of a nation as large as India  can be ensured only if appropriate safety measures and precautionary steps are taken right from the grassroots levels itself. Percolating down to the state level, government bodies have been eager to implement a public distribution system(PDS) early on in India. However, during the pandemic, the infrastructure of this system was not sufficient to satisfy the needs of the commoners living below the poverty threshold. With the loss of livelihoods and associated issues, malnutrition was on the rise along with heavy strain imposed on the existing healthcare system due to the infected patients. Perceiving this threat, the government of Kerala introduced an essential grocery food kit that could be collected from nearby ration shops. However, the contents of those food kits were a random assortment of edible items selected by the government without taking into consideration the preferences of its consumers. Hence, this paper sought to carry out a comprehensive analysis of various clustering algorithms to group different consumers into clusters and allocate a kit design based on their personal preferences. Among the different clustering algorithms, SVD-based clustering showed the most promising result, owing to the ease of dimensionality-reduction associated with it. This framework was implemented in a sampled consumer group, and future scope could extend to clustering based on a larger feature space taking into consideration factors such as geographical location, crop distribution, etc. Through this, one can infer that it is possible to develop such a personalized food distribution system, satisfying the consumers at the same time without leading to any food wastage.

\section{Acknowledgement}
This work was initially implemented at REBOOT Kerala Hackathon 2020, Finals organized by the Additional Skill Acquisition Program (ASAP), Kerala, and the Department of Higher Education, Government of Kerala. Later on, a detailed dataset of consumer preferences was collected by surveying a sampled population in Thiruvananthapuram, Kerala, India. which helped improve the findings. The authors are indebted to Dr. Ajeesh Ramanajun from the Computer Science Department of the College of Engineering Trivandrum and Prof. Tony S, from the Department of Mathematics, College of Engineering Trivandrum for their invaluable suggestions and guidance.

% Include references
\bibliography{References}

\end{document}